# Preliminary Development of a Wearable Device to Help Children with Unilateral Cerebral Palsy Increase Their Consciousness of Their Upper Extremity

Jade Clouatre[1,2], Alexandre Campeau-Lecours[1,2], Véronique Flamand[2,3]

[1]*Department of Mechanical Engineering, Université Laval, Quebec City, Canada,* [2]*Centre for Interdisciplinary Research in Rehabilitation and Social Integration, CIUSSS de la Capitale-Nationale, Quebec City, Canada,*

[3]*Department of Rehabilitation, Université Laval, Quebec City, Canada*

## ABSTRACT

Children with unilateral cerebral palsy have movement impairments that are predominant to one of their upper extremities (UE) and are prone to a phenomenon named "developmental disregard", which is characterized by the neglect of their most affected UE because of their altered perception or consciousness of this limb. This can cause them not to use their most affected hand to its full capacity in their day-to-day life. This paper presents a prototype of a wearable technology with the appearance of a smartwatch, which delivers haptic feedback to remind children with unilateral cerebral palsy to use their most affected limb, and which increase sensory afferents to possibly influence brain plasticity. The prototype consists of an accelerometer, a vibration motor and a microcontroller with an algorithm that detects movement of the limb. After a given period of inactivity, the watch starts vibrating to alert the user.

## INTRODUCTION

Children with unilateral cerebral palsy have movement impairments that are predominant to one of their upper extremities (UE) and that are not only caused by motor deficits, but by sensory deficits as well (e.g., in proprioception and stereognosis) [1,2,3]. More precisely, these children are prone to a phenomenon named "developmental disregard", which is characterized by the neglect of their most affected UE because of their altered perception or consciousness of this limb [4]. This can cause them not to use their most affected hand to its full capacity in their day-to-day life. However, an increased spontaneous use of their hand could ease the realization of many activities of daily living (ADL), as a majority of ADLs require the use of both hands in a coordinated manner [5].

It is thus relevant to develop innovative ways of targeting this reduced use of their most affected hand in the day-to-day life outside of conventional rehabilitation interventions. To this end, we aimed at developing a new wearable technology to help increase children's consciousness of their upper extremity. This wearable technology aims to provide vibratory feedback to the most affected UE of the user when it has been motionless for a certain amount of time. This technology can be worn in the natural environments of the children (thus over a prolonged period of time compared to clinical environment) to 1) increase the quantity of movements performed with the most affected UE because of the haptic feedback vibrations that remind the child to move after a period of inactivity and 2) to increase sensory afferents from the most affected UE, as vibration of a tendon or muscle can activate sensory receptors and pathways, and can possibly influence cortical motor excitability [6]. Such increased sensory stimulation is susceptible to enhance cerebral plasticity and lead to changes in the way that the sensorimotor systems control movement of the most affected UE. In order to ensure acceptability, this technology is designed with the appearance of a "smartwatch," which has a playful aspect that will help children to wear it in their day-to-day life. This paper presents a first prototype towards that goal. The hardware is first presented, followed by the algorithm, along with the interaction between the user and the watch.

## OBJECTIVES

The general aims of this project are 1) to develop a low-cost accessible wearable device delivering haptic feedback to users when their limb is inactive for a given time, 2) to validate the concept to see if it functionally helps children, and 3) to understand and characterize the parameters (e.g., type of vibration, inactivity time threshold) that leads to a better impact. This paper focuses solely on the first objective with the design of a first prototype to validate the concept and help guide the development. The prototype is developed through an iterative process, in collaboration with researchers in engineering and rehabilitation as well as occupational therapists, with a user-centered approach based on Design Thinking [7].

## DEVELOPMENT

### Hardware

The prototype consists of a microcontroller (Arduino Mega 2560), an inertial measurement unit (IMU) (3 accelerometers and 3 gyroscopes) (LSM9DS1), a DC vibration motor (Uxcell a16011800ux1186 DC 3V 12000 RPM), buttons and light-emitting diodes (LEDs), which are presented on Figure 1. The user arm acceleration (X, Y, Z) is obtained through the three axis accelerometer. An algorithm (detailed in the next section) is programmed on the microcontroller to robustly detect the user movements with that information and make the motor vibrate if the arm is inactive for a given time.

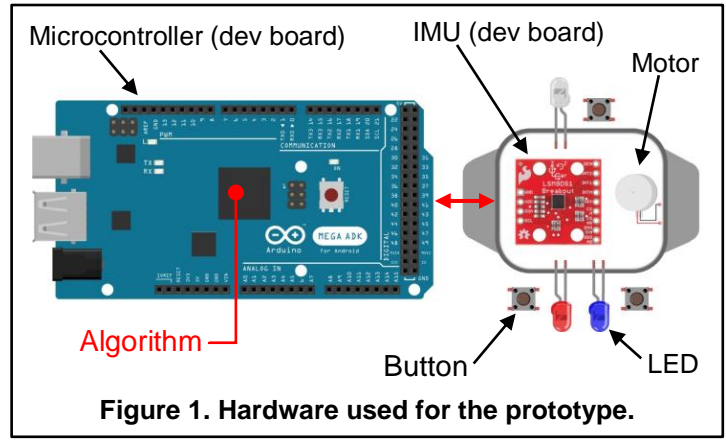

**Figure 1. Hardware used for the prototype.**

### Algorithm

Raw data acceleration (X, Y, Z), measured by the accelerometer, is acquired at 100 Hz and processed to transform the raw acceleration data into activity counts [8]. The aim of this algorithm is to robustly detect voluntary user movement of a high enough amplitude. Indeed, the raw acceleration signal can be affected by noise or impacts (e.g., vibrations when moving in a wheelchair, small arm movement) and such movement should not be considered as movements. Figure 2 shows the algorithm scheme. The raw acceleration is first sent to a band-pass filter (2nd order Butterworth filter with cutoff frequencies at 0.305 Hz and 1.615 Hz); this filter was adapted from Brønd, J. C., et al., who aimed to replicate the ActiLife[1] filter from ActiGraph. Thresholds are then applied on the filtered acceleration at 0.068 g and 2.13 g [8]. Then, the signal is converted from acceleration to counts, using the conversion of 0.01664 g/sec/count [9]. The counts are then integrated to epochs of length of 1 second. This process is applied independently on the X, Y and Z accelerations. Finally, the Euclidian norm is applied on the resulting sums to get the vector magnitude (VM) counts. This means that a new VM count is obtained at each 0.01 s (100 Hz) and includes the movement from the last second.

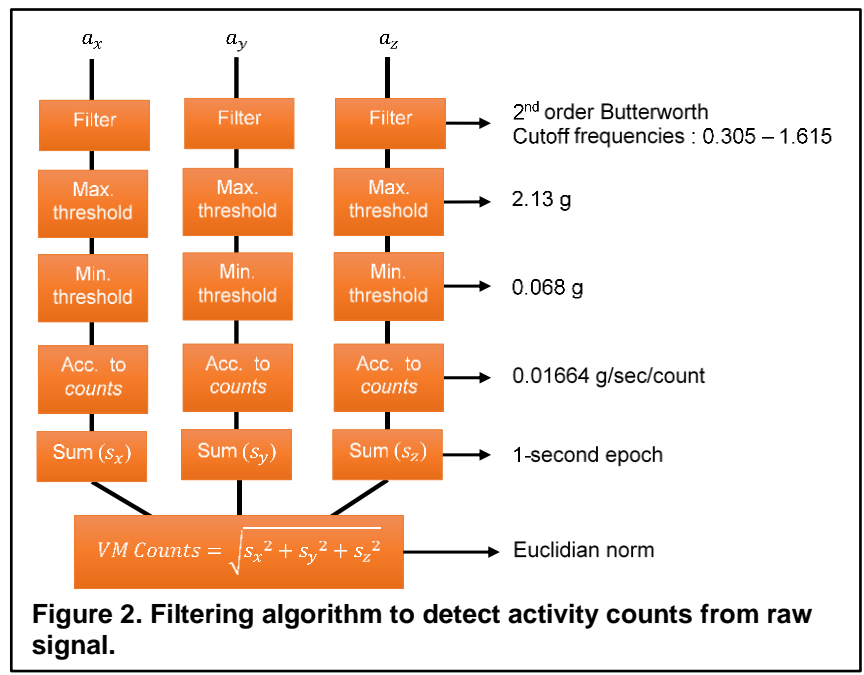

**Figure 2. Filtering algorithm to detect activity counts from raw signal.**

When the watch starts, an inactivity timer is set. If at a given time, the VM count exceeds a given threshold value, the device considers that the user has moved and the timer is reset; otherwise the timer keeps incrementing. The threshold value has been experimentally set at 125 (if the value is too low, the inactivity timer will reset for any small movement/vibration, and if it is too high it will require too large movements). This value can be changed and its optimal value will be further explored in the future. If no movement is detected after a given amount of time, the watch starts to vibrate for a few seconds to alert the user. The user can also stop the vibrations by moving his/her limb while wearing the watch. In addition to the threshold value, the inactivity time (10 s in the following example) and the vibration time (5 s in the following example) can be modified for each user. Figure 3 presents an example where a movement is detected (counts > 125 at approximate time = 7 s) and where the timer is reset, followed by 10 s of inactivity leading to vibrations (represented by the orange cross) for 5 s, and a timer reset. One should note that the accelerations generated by the vibration motor, which are of high frequency, do not translate into counts since they are filtered by the filtering algorithm (Figure 2).

---
[1] https://www.actigraphcorp.com/support/software/actilife/

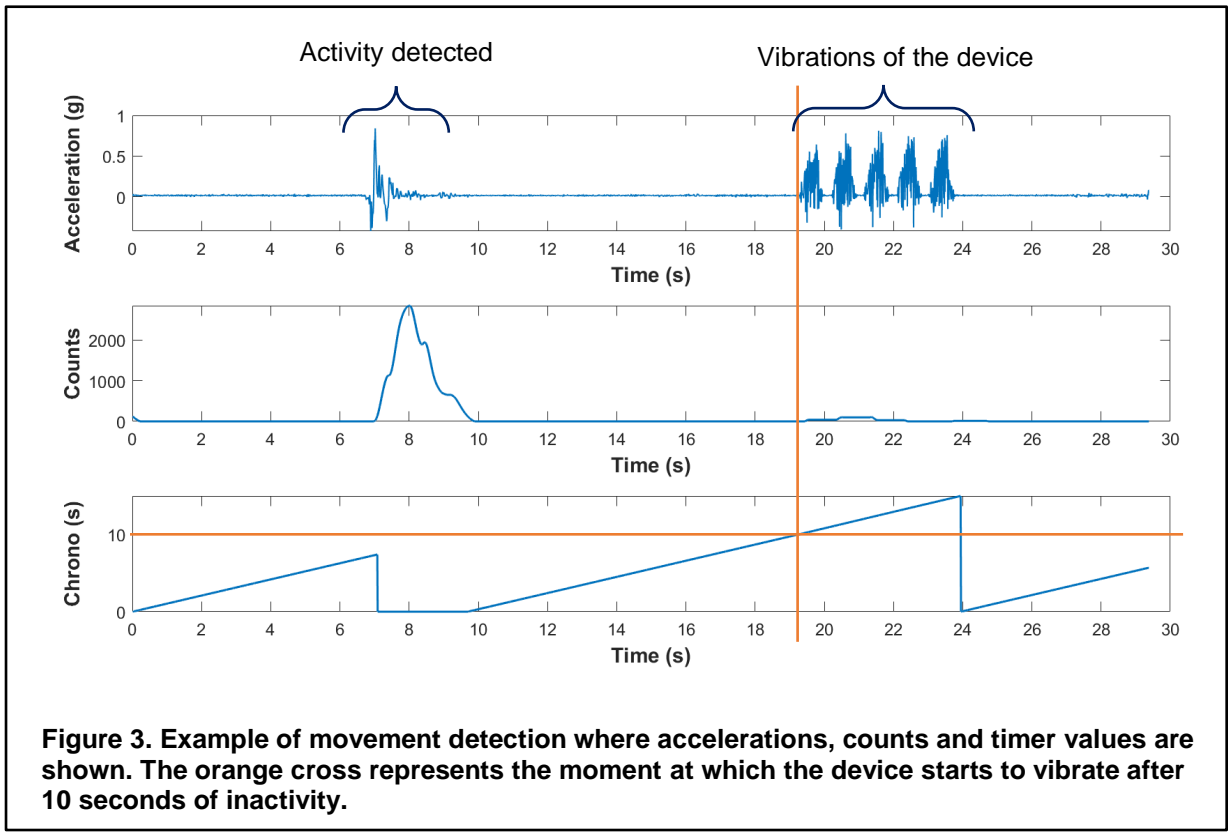

**Figure 3. Example of movement detection where accelerations, counts and timer values are shown. The orange cross represents the moment at which the device starts to vibrate after 10 seconds of inactivity.**

Interaction

The prototype also includes three buttons and three light-emitting diodes that enable the user to change options. The buttons are used to select between three different inactivity time lengths (specifically programmed for the user beforehand), activate a red LED that flashes in conjunction with the watch vibrations, and turn the watch off. A white LED stays on to indicate that the device is on, and a blue LED flashes once, twice or thrice to indicate which inactivity time length is selected after pressing the appropriate button.

## DISCUSSION

In this paper, a first prototype of a wearable device was developed to detect limb inactivity. An accelerometer was used to detect a limb acceleration, and an algorithm was proposed to robustly detect the absence of movement and alert the user through haptic feedback when the limb is inactive for a predetermined amount of time. The aim is to help children with unilateral cerebral palsy to increase their consciousness of their upper extremity and to move and use their most affected UE more often, while possibly influencing cortical motor excitability. This prototype has allowed to develop and refine the algorithms, to validate the concept with engineers and occupational therapists, and to validate the necessary hardware (e.g., number of buttons and LEDs). The main limitation of this prototype is that it is cumbersome and thus not usable in practice. Now that the required hardware has been validated, future work will consist in developing a miniature version of the device and explore the possibility to integrate the algorithms to an application on a smartwatch (e.g., Apple Watch). Finally, the prototype will be iteratively defined and validated in an ecological situation with children living with unilateral cerebral palsy.

## CONCLUSION

This paper focussed on the design of a low-cost accessible wearable device able to provide haptic feedback to users with unilateral cerebral palsy when their most affected limb is inactive for a given amount of time. Future work will consist in developing a compact version to validate the concept and see if it helps children, in validating the concept to see if it functionally helps children, and in understanding and characterizing the parameters (e.g., type of vibration, inactivity time) and their influence on the device performance.

**ACKNOWLEDGEMENTS**

This work was supported by the Fonds de recherche Québec Nature et Technologies (FRQNT - INTER Strategic Network [Engineering of Interactive Rehabilitation Technologies / Ingénierie de technologies interactives en réadaptation]) under grant 2020-RS4-265381.

**REFERENCES**

[1] Gordon AM, Bleyenheuft Y, Steenbergen B. Pathophysiology of impaired hand function in children with unilateral cerebral palsy. *Dev Med Child Neurol* 55: 32–37, 2013.

[2] Goble DJ, Hurvitz EA, Brown SH. Deficits in the ability to use proprioceptive feedback in children with hemiplegic cerebral palsy. *Int J Rehabil Res Int Zeitschrift fur Rehabil Rev Int Rech Readapt* 32: 267–269, 2009.

[3] Van Heest AE, House J, Putnam M. Sensibility deficiencies in the hands of children with spastic hemiplegia. *J Hand Surg Am* 18: 278–281, 1993.

[4] Taub E. Parallels between use of constraint-induced movement therapy to treat neurological motor disorders and amblyopia training. *Dev Psychobiol* 54: 274–92, 2012.

[5] Krumlinde-Sundholm L. On the other hand: About successful use of two hands together [Online]. *Dev Med Child Neurol* 51: 39, 2009. http://www.embase.com/search/results?subaction=viewrecord&from=export&id=L70198940.

[6] Annino G, Alashram AR, Alghwiri AA, Romagnoli C, Messina G, Tancredi V, Padua E, Mercuri NB. Effect of segmental muscle vibration on upper extremity functional ability poststroke: A randomized controlled trial. Medicine (Baltimore) 98: e14444, 2019.

[7] Brown, T. (2009). Change by design, 1-5

[8] Brønd, J. C., Andersen, L. B., Arvidsson, D. Generating actiGraph counts from raw acceleration recorded by an alternative monitor. *Medicine & Science in Sports & Exercise, 49,* 2351-2360, 2017

[9] ActiGraph (2018, November 8). *What are counts*, Retrieved January 10, 2020 from https://actigraphcorp.force.com/support/s/article/What-are-counts